\documentclass{article}

    \PassOptionsToPackage{numbers, compress}{natbib}

\usepackage[preprint]{neurips_2024}

\usepackage[dvipsnames]{xcolor}         
\definecolor{linkColor}{rgb}{0.2,0.4,0.6}
\usepackage[utf8]{inputenc} 
\usepackage[T1]{fontenc}    
\usepackage[colorlinks=true,linkcolor=linkColor,citecolor=linkColor,filecolor=linkColor,urlcolor=linkColor]{hyperref}       
\usepackage{url}            
\usepackage{booktabs}       
\usepackage{amsfonts}       
\usepackage{nicefrac}       
\usepackage{microtype}      

\usepackage{graphicx}
\usepackage{arydshln}

\usepackage{tabularx} 
\usepackage{booktabs} 

\usepackage{booktabs}
\usepackage{multirow}
\usepackage{caption}
\usepackage{subcaption}
\usepackage{makecell}
\usepackage{csquotes}
\usepackage{epigraph}
\usepackage{nicematrix}
\usepackage{pgfplots}
\usepackage{tikz}
\usepackage[algo2e, ruled,linesnumbered]{algorithm2e}
\usepackage{graphicx}
\usepackage{colortbl}
\usepackage{tcolorbox}
\usepackage{wrapfig}
\usepackage{floatrow}
\usepackage{lipsum}
\usepackage{tablefootnote}
\usepackage{threeparttable}
\newfloatcommand{capbtabbox}{table}[][\FBwidth]

\pgfplotsset{compat=1.3}

\RequirePackage{algorithm}
\RequirePackage{algorithmic}

\usepackage{multirow}
\usepackage{amsmath}
\usepackage{capt-of}
\usepackage{tabularx}
\usepackage{epsfig}
\usepackage{amssymb}
\usepackage{amsfonts}
\usepackage{booktabs}
\usepackage{scalerel}
\usepackage[inline]{enumitem}
\usepackage{listings}
\usepackage{varwidth}
\usepackage[export]{adjustbox}
\usepackage{tikz}
\usetikzlibrary{tikzmark}

\usepackage{stmaryrd}
\usepackage{bbm}
\usepackage{wrapfig}
\usepackage{pifont}
\usepackage[noabbrev]{cleveref}

\definecolor{deepblue}{rgb}{0,0,0.5}
\definecolor{officeblue}{RGB}{0,102,204}
\definecolor{deepred}{rgb}{0.6,0,0}
\definecolor{deepgreen}{rgb}{0,0.5,0}
\definecolor{mybrickred}{RGB}{182,50,28}

\definecolor{fillcolor}{RGB}{216,217,252}

\usepackage{etoolbox}
\usepackage{framed}

\newif\ifxetexorluatex
\ifxetex
  \xetexorluatextrue
\else
  \ifluatex
    \xetexorluatextrue
  \else
    \xetexorluatexfalse
  \fi
\fi
\newcommand*\quotesize{60} 
\newcommand*{\openquote}
   {\tikz[remember picture,overlay,xshift=-4ex,yshift=-2.5ex]
   \node (OQ) {\fontsize{\quotesize}{\quotesize}\selectfont``};\kern0pt}

\newcommand*{\closequote}[1]
  {\tikz[remember picture,overlay,xshift=4ex,yshift={#1}]
   \node (CQ) {\fontsize{\quotesize}{\quotesize}\selectfont''};}

\colorlet{shadecolor}{white}

\newcommand*\shadedauthorformat{\emph} 

\newcommand*\authoralign[1]{%
  \if#1l
    \def\authorfill{}\def\quotefill{\hfill}
  \else
    \if#1r
      \def\authorfill{\hfill}\def\quotefill{}
    \else
      \if#1c
        \gdef\authorfill{\hfill}\def\quotefill{\hfill}
      \else\typeout{Invalid option}
      \fi
    \fi
  \fi}
{\authoralign{#1}
\ifblank{#2}
   {\def\shadequoteauthor{}\def\yshift{-2ex}\def\quotefill{\hfill}}
   {\def\shadequoteauthor{\par\authorfill\shadedauthorformat{#2}}\def\yshift{2ex}}
\begin{snugshade}\begin{quote}\openquote}
{\shadequoteauthor\quotefill\closequote{\yshift}\end{quote}\end{snugshade}}

\newfloat{Code}{htbp}{Code}

\usepackage{amsmath,amsfonts,bm}

\def\eqref#1{equation~(\ref{#1})}

\def\1{\bm{1}}

\DeclareMathAlphabet{\mathsfit}{\encodingdefault}{\sfdefault}{m}{sl}
\SetMathAlphabet{\mathsfit}{bold}{\encodingdefault}{\sfdefault}{bx}{n}

\newcommand{\rparagraph}[1]{\vspace{1.2mm}\noindent\textbf{#1.}}
\newcommand{\iparagraph}[1]{\vspace{1.2mm}\noindent\textit{#1.}}

\newcommand{\iparagraphnodot}[1]{\vspace{0.0mm}\noindent\textit{#1}}

\newcommand{\benchmark}{\textsc{11Plus-Bench}\xspace}

\title{\benchmark: Demystifying Multimodal LLM \\Spatial Reasoning with Cognitive-Inspired Analysis}

\newcommand{\ltl}{\texttt{[LTL]}}
\newcommand{\msr}{\texttt{[MSR]}}
\newcommand{\casia}{\texttt{[CAS]}}
\newcommand{\onc}{\texttt{[ONC]}}
\newcommand{\cfi}{\texttt{[CFI]}}
\newcommand{\val}{\texttt{[VAL]}}

\author{
Chengzu Li\textsuperscript{\texttt{[MSR\thanks{Work done during internship at Microsoft Research.} ,LTL]}}
\quad 
Wenshan Wu\textsuperscript{\msr}
\quad
Huanyu Zhang\textsuperscript{\texttt{[MSR\textsuperscript{*},CAS]}}
\\
\textbf{Qingtao Li\textsuperscript{\msr}} \quad
\textbf{Zeyu Gao\textsuperscript{\onc}}
\quad
\textbf{Yan Xia\textsuperscript{\msr}} \\
\textbf{José Hernández-Orallo\textsuperscript{\texttt{[CFI,VAL]}}}
\quad
\textbf{Ivan Vuli{\'c}\textsuperscript{\ltl}}
\quad
\textbf{Furu Wei\textsuperscript{\msr}}
 \vspace{1mm}
 \\
 {\href{https://aka.ms/GeneralAI}{https://aka.ms/GeneralAI}} \\
 \\
   \textsuperscript{\msr}Microsoft Research~~
   \textsuperscript{\ltl}Language Technology Lab, University of Cambridge\\
   \textsuperscript{\casia}Institute of Automation, Chinese Academy of Sciences\quad \\
  \textsuperscript{\onc}Department of Oncology, University of Cambridge \\
    \textsuperscript{\cfi}Leverhulme Centre for the Future of Intelligence, University of Cambridge \\
  \textsuperscript{\val}VRAIN, Universitat Politècnica de València
}

\begin{document}
\maketitle

\begin{abstract}
For human cognitive process, spatial reasoning and perception are closely entangled, yet the nature of this interplay remains underexplored in the evaluation of multimodal large language models (MLLMs). 
While recent MLLM advancements show impressive performance on reasoning, their capacity for human-like spatial cognition remains an open question.
In this work, we introduce a systematic evaluation framework to assess the spatial reasoning abilities of state-of-the-art MLLMs relative to human performance. 
Central to our work is \benchmark, a high-quality benchmark derived from realistic standardized spatial aptitude tests. 
\benchmark also features fine-grained expert annotations of both perceptual complexity and reasoning process, enabling detailed instance-level analysis of model behavior.
Through extensive experiments across 14 MLLMs and human evaluation, we find that current MLLMs exhibit early signs of spatial cognition.
Despite a large performance gap compared to humans, MLLMs' cognitive profiles resemble those of humans in that cognitive effort correlates strongly with reasoning-related complexity. 
However, instance-level performance in MLLMs remains largely random,
whereas human correctness is highly predictable and shaped by abstract pattern complexity.
These findings highlight both emerging capabilities and limitations in current MLLMs’ spatial reasoning capabilities and provide actionable insights for advancing model design.
\end{abstract}

\section{Introduction}

Many achievements of Large Language Models (LLMs) \cite{brown2020language, ouyang2022training, anil2023palm} and their multimodal variants (MLLMs) \cite{hurst2024gpt, gemini2024, gemini2025thinking} are largely concentrated in domains where reasoning can be framed through symbolic sequence processing, including code generation \cite{austin2021program,lai2023ds}, mathematical problem solving \cite{lu2024mathvista,wang2024measuring,wang2025mathcodervl}, and question answering \cite{yang-etal-2015-wikiqa,hendrycks2020measuring,li-etal-2025-large-language-models,zhangmme}.
Human intelligence goes beyond symbolic processing. 
It relies heavily on perceptual intuition and mental imagery to simulate hypothetical scenarios via object-based imagery (e.g., of shapes) and spatial imagery (e.g., of locations) \cite{moulton2009imagining, kozhevnikov2005spatial}, which is still underexplored with MLLMs \cite{li2025imagine,xu2025visual}. 
Spatial reasoning, also referred to spatial intelligence in cognitive science, encompasses all thinking about spatial content: object shape or location, and manipulating, imagining, or inferring relationships between objects in space \cite{Newcombe2024Spatial}. 
Carroll's Three-Stratum Theory of Intelligence \cite{Carroll1993, carroll1997three} places \textit{Visualization} and \textit{Spatial Relations} as core narrow abilities within the general spatial intelligence domain (Gv), contributing to general intelligence (\textit{g}) as evidenced by empirical research \citep{deary2010neuroscience}.
Spatial reasoning is crucial for success in STEM fields, visuospatial memory, navigation, and mechanical reasoning \cite{Harvey01061985, Wai2009SpatialAF, fa708fa9705b4671b137d87ae44d25ff, li-etal-2024-semantic, Zhou20241220}. 
Despite its fundamental importance to human intelligence, spatial reasoning remains a relatively underexplored area in the evaluation of artificial intelligence.

\begin{figure*}[t]
    \centering
    \includegraphics[trim={0cm 3.7cm 0cm 4.5cm}, clip, width=\textwidth]{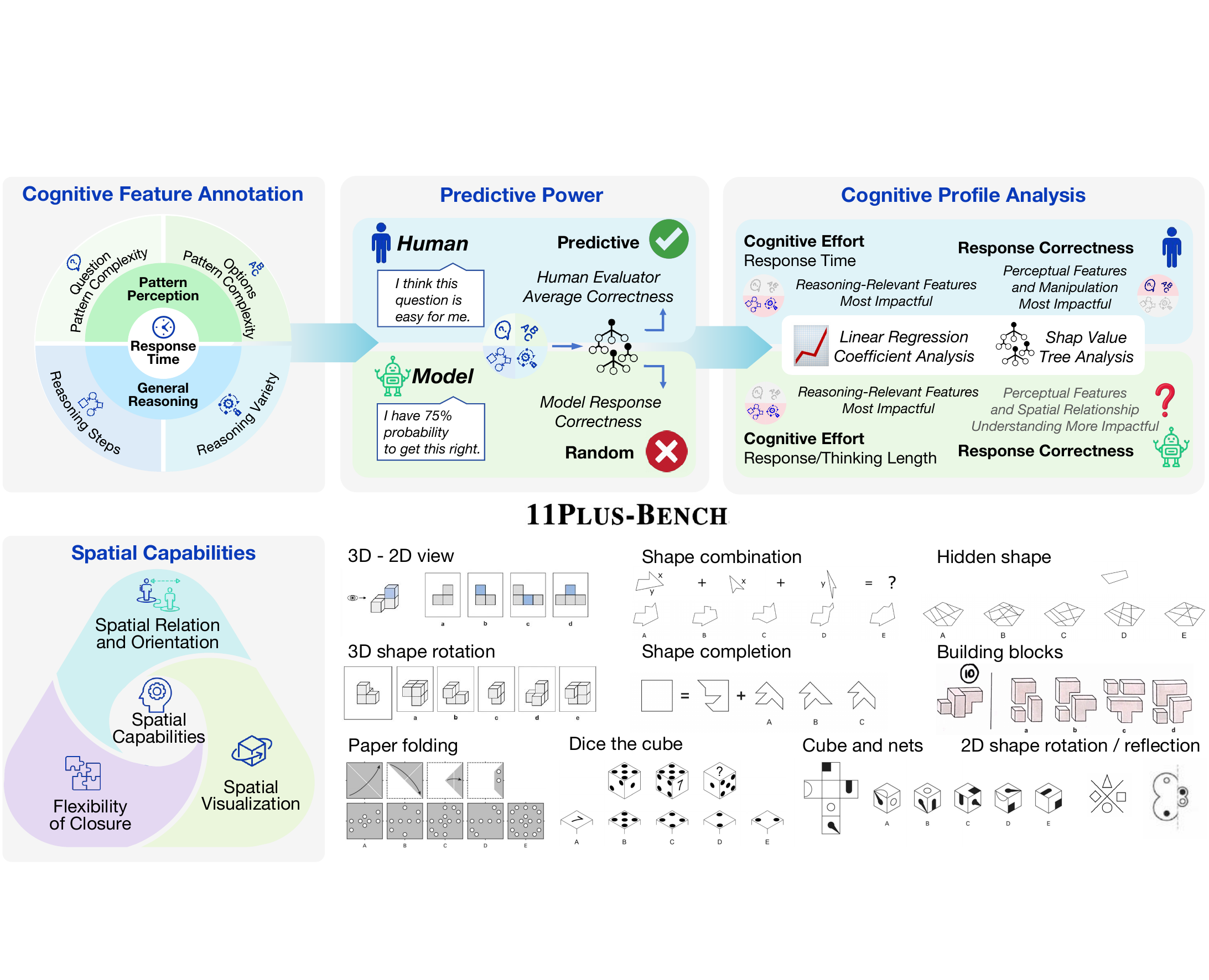}
    \caption{\textbf{Overview of evaluation framework with \benchmark}, including fine-grained annotations of cognitive features across diverse tasks targeting three core spatial capabilities. These annotations enable predictive modeling of correctness for both humans and MLLMs, followed by cognitive profile analysis to identify key features that influence accuracy and cognitive load.}
    \label{fig:evaluation framework}
\end{figure*}

Existing work evaluating MLLM spatial reasoning has largely relied on aggregate metrics such as overall or task-wise accuracy \cite{Ramakrishnan2025,MindTheGap2025,xu2025defining}, which offers only a coarse view of model ability. 
These holistic evaluations often conflate distinct cognitive processes, such as perception, symbolic reasoning, and spatial inference \cite{zhou2025general}, limiting interpretability and obscuring a model's true capabilities in spatial reasoning.
Consequently, pinpointing specific skill deficits in current systems from aggregated metrics is challenging, leading to potential misattributions (e.g., mistaking perceptual failures for reasoning deficits \citep{chollet2024arc,chollet2025arc}) and hindering clear improvement pathways for MLLM spatial cognition. 
Furthermore, despite referencing human cognitive tests as testbed, comparisons between human cognition and model behavior in existing work remain relatively shallow \cite{xu2025defining,wei2025position, zhang2025scaling}, failing to specifically highlight current MLLM systems' deficiencies compared to human capabilities. 

To address these gaps, we ask: Do current MLLMs engage in spatial reasoning in a manner aligned with human cognition?
We refer to the strategies and capabilities of perception, interpretation, and reasoning in spatial contexts as the model's cognitive profile, and we aim to facilitate a parallel comparison of these cognitive profiles between humans and MLLMs.

To this end, we present this evaluation framework centered around \benchmark, a newly-introduced high-quality benchmark grounded in standardized spatial aptitude tests used in human cognitive assessments \cite{Uttal2013, spatialcogandsciachievement}. 
This design isolates spatial reasoning from confounding factors such as commonsense knowledge or numerical ability.
Unlike traditional benchmarks that emphasize aggregate accuracy, \benchmark supports \textbf{instance-level comparisons} between the correctness of model responses and the perceived difficulty of human behaviors. 
It 
features \textbf{fine-grained expert annotations} of \textbf{cognitive features}, capturing both visual pattern complexity (perceptual load) and reasoning process (inference difficulty), allowing us to investigate and disentangle different factors that influence system behavior.
To compare with human performance, we conduct human evaluations with three participants and use response time as a proxy for cognitive load \cite{barrouillet2007time, jintelligence4040014}. 
Our annotations exhibit high inter-annotator agreement and strong predictive power for participant response time with annotated cognitive features, validating the benchmark’s quality and interpretability.
\benchmark also minimizes contamination concerns by collecting expert annotations for data with no golden answers (over 50\%) and holding out a test split composed of problems sourced from commercial test providers that are not publicly available.

Experimental results across 14 state-of-the-art MLLMs reveal a substantial performance gap between models and humans, emphasizing the current limitations of MLLMs in spatial reasoning.
While advanced proprietary MLLMs show early signs of spatial reasoning ability, their instance-level performance remains random and 
poorly predictable with human-inspired cognitive features above.
Further analysis uncovers both convergence and divergence in cognitive profiles. 
Reasoning-related complexity correlates  strongly with cognitive load, measured by response time in humans and token counts of response for MLLMs.
However, model performance is more sensitive to understanding low-level visual cues such as image resolution and spatial relations, whereas human accuracy is primarily influenced by abstract pattern complexity.
This blend of similarity and divergence reveals both the emergence of spatial reasoning capabilities in MLLMs and their current deficiencies.
Unlike humans, whose spatial reasoning is structured, MLLMs often lack the robustness and compositional understanding necessary for consistent, human-like spatial cognition.
\section{Related Work}

\paragraph{Spatial Aptitude Test in Cognitive Science}
Human spatial ability includes \textit{intrinsic} object-centred skills (e.g., mental rotation, paper‑folding) and \textit{extrinsic} environment‑centred skills (e.g., perspective taking, navigation) \citep{hodgkiss2018spatial}. 
Classic experimental work on mental rotation by \citet{Shepard1971} and \citet{Cooper1975} frames rotation as a continuous internal transformation. 
Factor‑analytic syntheses later showed that rotation loads on a separable spatial factor distinct from verbal or numerical reasoning \citep{McGee1979,Linn1985,Carroll1993}. 
Perspective‑taking studies, notably \citet{Hegarty2004}, demonstrated a double dissociation from mental rotation, motivating multi‑dimensional test batteries such as the Vandenberg–Kuse Mental Rotation Test, Paper‑Folding and Spatial Orientation tests \citep{Ekstrom1976}.  
Training meta‑analyses confirm that spatial skills are malleable and transfer to STEM success \citep{Cheng2014,Uttal2013}. 
Neuropsychological reviews link these competencies to parietal–frontal circuits and hippocampal place/grid coding, underscoring their foundational role in cognition \citep{Burgess2008,Husain2007}. Together, these findings provide both theoretical structure and validated psychometrics that any AI‐oriented spatial benchmark should respect.

\paragraph{Spatial Cognition with MLLMs}
Early multimodal benchmarks such as CLEVR \citep{Johnson2017} and NLVR$\,2$ \citep{Suhr2019} introduced synthetic and natural‑image tasks that hinge on recognising static binary relations (e.g., \textit{left of, behind}). 
Subsequent datasets, e.g. SpatialSense \citep{Yang2019}, Spatial‑MM \citep{Shiri2024}, and Comsa \& Narayanan’s preposition suite \citep{Comsa2023}, tightened the focus on fine‑grained relational semantics. 
Yet performance plateaus suggest that current MLLMs still rely on language priors rather than genuine geometric reasoning \citep{Wang2024,xu2025visual}.  
Dynamic extensions (CLEVRER \citep{Yi2019}, TopViewRS \citep{li-etal-2024-topviewrs}, VSI‑Bench \citep{Yang2024}) add temporal sequences, but typically restrict transformations to planar translation or simple collisions, leaving rotation, reflection, and multi‑step composite reasoning under‑explored.  
Holistic test batteries such as \textit{MindtheGap} \citep{MindTheGap2025} and SAT \citep{Ray2024} broaden the coverage by emulating psychometric tasks. 
Despite the breadth, analyses remain largely descriptive, reporting that “MLLMs fail” without isolating \textit{why} (e.g., frame‑of‑reference confusion, object‑correspondence errors) or benchmarking against human baselines \citep{Ramakrishnan2025}.
Our benchmark, \benchmark, adopts a cognitive science–informed taxonomy and includes human performance statistics for each item, enabling detailed, parallel analysis of model and human cognitive profiles.
\section{\benchmark Benchmark}

\subsection{Collection of Tasks}
\label{subsec:task scope}
\rparagraph{Spatial Capabilities}
Human cognitive development involves several key capabilities that collectively form spatial intelligence. 
Psychometric research has identified and quantified these through standardized tests, capturing dimensions such as Spatial Relation and Orientation, Spatial Visualization, Flexibility of Closure, Perceptual Speed, Spatial Memory, and more \citep{Shepard1971, Linn1985, Burgess2008, yilmaz2009development, johnson2017predictive, wei2020psychometric, ekstrom2023spatial}.

However, not all these capabilities are equally relevant for evaluating current MLLMs, given fundamental differences in reasoning mechanisms between human cognition and machine learning models. 
For instance, perceptual speed is less critical for current MLLM paradigms, which do not process information in real-time like humans. 
Similarly, factors like spatial memory \citep{Burgess2008,ekstrom2023spatial} (e.g., recalling routes or locations over time) or kinesthetic spatial reasoning (understanding space through bodily movement) \citep{presson1987orientation,proske2009kinaesthetic} may not directly translate to current MLLM architectures which primarily operate on simulated static multimodal inputs.
Therefore, we select three representative spatial capabilities:
\begin{itemize}[leftmargin=*]
    \item \textit{Spatial Relation and Orientation (\textbf{SRO})}: Involves understanding relationships between objects in space, including distance, direction, and position \citep{newcombe2005development,yilmaz2009development}. It is essential for tasks requiring recognition of spatial configurations and interrelations.
    \item \textit{Spatial Visualization} (\textit{\textbf{SV}}): Refers to the ability to mentally manipulate and transform spatial information \citep{michael1957description,Shepard1971}. This is important for tasks involving mental rotation, pattern recognition, and imagining as well as manipulating objects or scenes.
    \item \textit{Flexibility of Closure} (\textit{\textbf{FoC}}): Pertains to the ability to perceive and mentally complete incomplete patterns or shapes \citep{yilmaz2009development}. This cognitive ability is crucial for solving problems that require identification of missing or occluded elements.
\end{itemize}

\rparagraph{Task Selection}
We utilize well-established psychometric tests corresponding to the selected capabilities \citep{harris2013new, lovett2013modeling, jirout2015building, parkinson2018investigating, gunalp2019spatial, uttal2024can}. 
These tests are widely acknowledged and developed in cognitive science, ensuring a fair and parallel comparison between AI systems and humans. 
Because most psychometric tests use diagrams and structured questions as multimodal input, they also allow for controlled experiments in terms of task complexity while controlling other irrelevant factors to spatial intelligence, such as entity recognition in real-world images.
Table \ref{apptab:capability and task} presents the correspondence between tasks and capabilities, and Figure \ref{fig:evaluation framework} provides concrete examples. 
See Appendix \ref{appsubsec:benchmark construction} for detailed definitions of each task.

\subsection{Collection of Cognitive Features}
\label{subsec:confoundingfactors}

Answering spatial cognition questions not only requires spatial reasoning but also depends on visual perception and general reasoning performance. 
These factors influence the probability of a correct response from both humans and machines but do not directly measure spatial reasoning. 
For a fine-grained explainable investigation, we collect performance-relevant \textit{cognitive features} as follows:

\rparagraph{Visual Perception}
More complex patterns require greater cognitive load for humans to perceive and analyze. 
For both the question and options, we quantify pattern complexity as the number of atomic components in the patterns as key features, defined by how humans perceive and analyze patterns (details on the objective definition of `key features' can be found in Appendix \ref{appsubsec:benchmark construction}).

\rparagraph{General Reasoning}
Longer reasoning chains indicate greater question complexity and a higher likelihood of error \citep{garey2002computers, johnson2010mental}. 
Transitions among reasoning types, such as logical deduction and pattern recognition, add extra cognitive load. 
These features are distinct from intrinsic spatial cognition but influence reasoning time or response correctness. 
Variations in these features are subjectively profound, as different individuals may adopt different reasoning chains, especially for more complex questions. 
To account for this subjective variation, we annotate the general reasoning process by requiring human annotators to choose from four predefined categories of atomic operations: \textit{Pattern Matching}, \textit{Spatial Relation Analysis}, \textit{Spatial Manipulation}, and \textit{Logical Deduction}, each comprising a set of specific operations with details in Appendix \ref{appsubsec:benchmark construction}.

In addition to these cognitive features, \textit{bounding boxes} of question and option patterns are also collected in pixel coordinates.

\subsection{Benchmark Construction}
\label{subsec:benchmark construction}
To facilitate the evaluation framework, we construct the \benchmark with realistic cognitive science test targeted for teenagers aged 11 or above (\textsc{11Plus}). 
We compile the public portion of our benchmark by crawling the web using carefully chosen spatial reasoning keywords. 
A rule-based filtering pipeline is then introduced to discard irrelevant, ambiguous, or non-spatial reasoning samples, ensuring data quality and relevance. 
Implementation details are provided in Appendix~\ref{appsubsec:benchmark construction}.
Concurrently, the private portion of our benchmark is sourced by purchasing materials from official test centers. 
This dual approach, combining newly annotated public data with proprietary test-center materials, creates a robust and professional dataset that captures a broad spectrum of spatial cognition challenges while ensuring data quality and contamination control for model evaluation.

All annotations were performed by three human experts, who are postgraduate-level or higher with mathematical or engineering backgrounds.
Annotators were trained using standardized guidelines to ensure consistency and reliability across the dataset.
They annotated the entire public set and an additional 100 samples drawn from the private set, creating a diverse and robust foundation for evaluating spatial reasoning. 
Data examples deemed low-quality, without a correct answer, or not belonging to spatial cognition were manually filtered and discarded. 
By combining thorough filtering with expert human annotation, we ensure the benchmark reflects genuine spatial cognition challenges and minimizes errors. 

\begin{wrapfigure}{r}{0.62\textwidth}
    \vspace{-10pt}
    \includegraphics[trim={12.5cm 9cm 1.7cm 7cm}, clip, width=\textwidth]{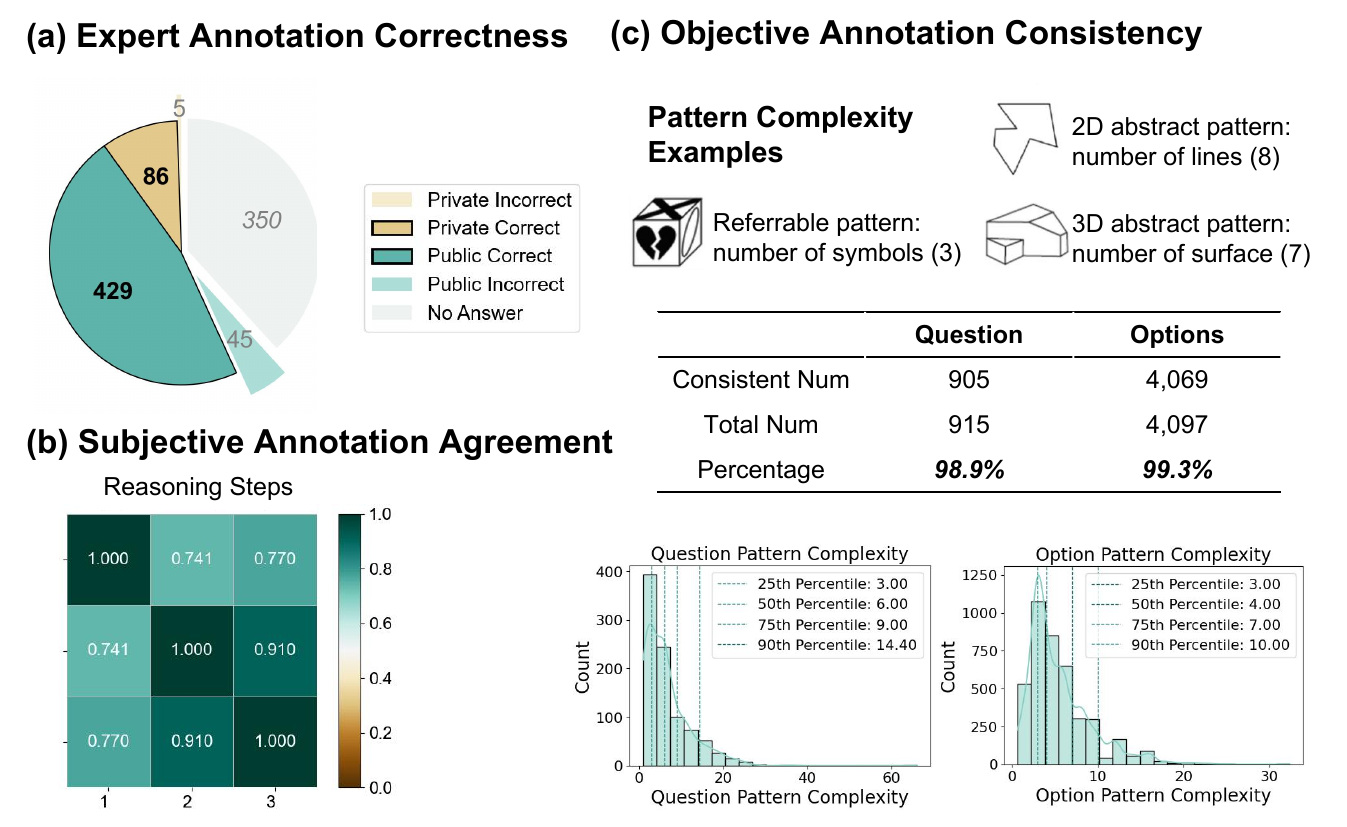}
    \caption{\textbf{Quality analysis of expert data collection.} Expert annotations achieve high accuracy on private data with golden answers and exhibit strong agreement across both subjective and objective annotation fields.}
    \label{fig:data collection}
    \vspace{-10pt}
\end{wrapfigure}

\paragraph{Benchmark Quality Analysis}
The fine-grained annotated benchmark contains 824 data points in the public set and 91 data points in the private set after filtering, all annotated by 3 domain experts.
The annotations exhibit strong internal consistency and correctness, underscoring the high quality of the dataset, as shown in Figure \ref{fig:data collection}.
The annotated answers achieve 94.5\% accuracy on private set against gold-standard labels.
For subjective fields such as Reasoning Steps, we observe a high level of annotator agreement, with Pearson correlation coefficients typically around or above 0.8. 
The objective pattern complexity for both questions and options shows perfect agreement among annotators, with numbers strictly aligned. 
Appendix \ref{appsubsec:benchmark construction} provides more information about our benchmark.

\paragraph{Data Highlights}
Here are the key highlights of \benchmark:
\begin{itemize}[leftmargin=*]
    \item \textbf{\textit{More Realistic Data}}: \benchmark contains two separate data splits (public with 824 examples \& private with 91 examples), all derived from realistic 11Plus spatial aptitude test. The public set was crawled from the web, while the private set was purchased from test centers and involves copyrights and intellectual properties.
    \item \textbf{\textit{Lower Risk of Data Contamination}}: With experts annotating over 50\% data with no golden answer available and withholding the private set due to intellectual property considerations, \benchmark significantly lowers the risk of data contamination when evaluating model performance.
    \item \textbf{\textit{Richer Cognitive Features}}: In addition to the golden answer, \benchmark provides richer fields including not only \textit{bounding boxes} for patterns but also \textit{visual perception} complexity, \textit{general reasoning} process as cognitive annotation. 
\end{itemize}

\section{Experiments and Results}
\label{sec:results}

\subsection{Experimental Setups}
\paragraph{Models}
To comprehensively assess the spatial cognition capabilities of contemporary Multimodal Large Language Models (MLLMs), we selected a diverse suite of 14 models. 
This selection encompasses both open-sourced and close-sourced architectures, varying significantly in their parameter counts and underlying designs. 
Specifically, we evaluated four open-sourced models: Qwen-VL-2.5 \citep{bai2025qwen2} (with 3B and 7B parameters) and Gemma 3 \citep{team2025gemma} (with 12B and 27B parameters). 
Complementing these, we included ten close-sourced MLLMs: GPT-4o, GPT 4.1 mini, GPT 4.1 nano, GPT-o1, GPT-o3, GPT-o4-mini, GPT4.1, Gemini 2.0 Flash preview, Gemini 2.5 Flash preview and Gemini 2.5 Pro preview \citep{hurst2024gpt, gpto1, gpt41, gpto3o4mini, gemini2024, gemini2025thinking}. 
This curated set allows for a broad analysis of how model scale and accessibility correlate with performance. 

\paragraph{Task Settings}

The evaluation methodology extends traditional Visual Question Answering (VQA) benchmarks by also presenting multiple images as options in response to a given question. 
We investigate two distinct presentation formats to evaluate the MLLMs' spatial cognition:
\begin{enumerate}
\item \textbf{Single Composite Image}: In this setup, a single image is presented to the model, as with humans. This image contains both the primary image relevant to the question and all candidate option images arranged spatially. This approach is adopted by previous works in benchmarking the spatial cognition performance of MLLMs \citep{MindTheGap2025,Ramakrishnan2025,xu2025defining}.
\item \textbf{Separate Images with Bounding Box Annotations}: The primary image and each option image are cropped from the original images as distinct, separate visual inputs. This allows models to potentially ground their reasoning more precisely on specific visual elements.
\end{enumerate}
The performance of the MLLMs across all tasks is quantified by their accuracy in selecting the correct option image that answers the posed question. 

\paragraph{Human Evaluation}
Three participants who are not involved in the annotation process are recruited in order to assess human performance on \benchmark benchmark, strictly adhering to ethical regulations. 
The examples for human evaluation are uniformly sampled from different tasks, with all data being used for specific task if the available examples are less than sampling requirements, resulting in 402 examples in total.
In addition to collecting participants’ selected answers, we record the \textit{response time} for each human participant to answer the question, measured in seconds, as an outcome-driven proxy for overall cognitive load \citep{barrouillet2007time,jintelligence4040014}.

\subsection{Results}
\label{subsec:discussion}

\begin{figure*}[t]
    \centering
    \includegraphics[trim={2.5cm 8.5cm 0.4cm 2.5
    cm}, clip, width=\textwidth]{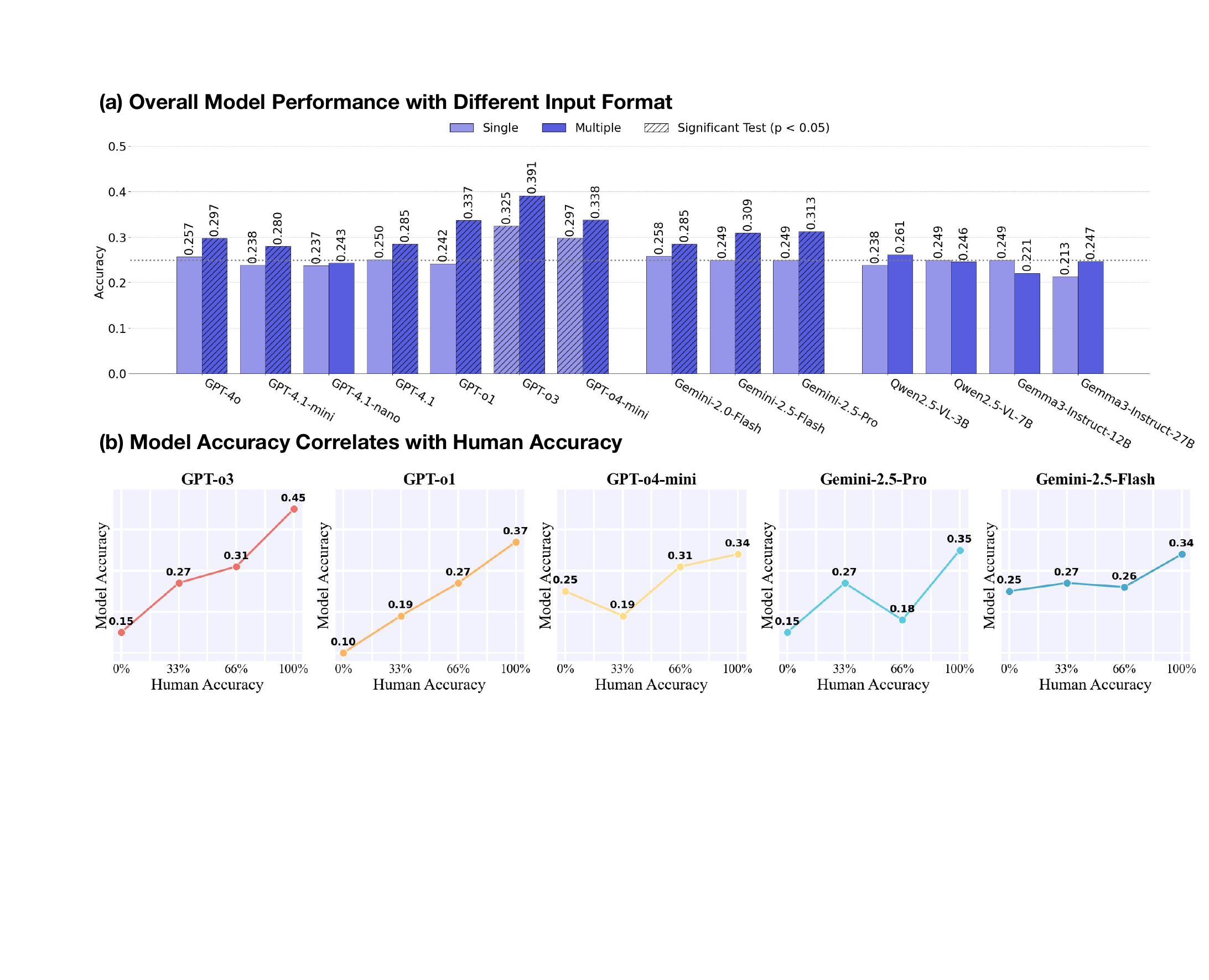}
    \caption{\textbf{(a)} Models perform better with multiple separate images as input compared to a single image. With multiple-image input, most closed-source models pass the significance test (p < 0.05) over random guess, whereas still all open-sourced models fail. \textbf{(b)}  MLLM performance correlates with human accuracy (0–3 correct responses across all participants), achieving higher accuracy on instances rated as easier by human evaluators. }
    \label{fig:model performance}
\end{figure*}

\paragraph{Human Performance}
Human participants achieve accuracies of 72\%, 87\% and 85\% across the 402 examples. 
Of all the examples, 241 of them are answered correctly by all three participants, 115 are answered correctly by two and 46 questions are answered correctly by one or none. 
Response times exhibit moderate correlation among participants, with a Pearson correlation coefficient exceeding 0.4. 
Additionally, the intraclass correlation coefficient ($ICC2=0.529$) indicates moderate agreement, and the average response time is deemed reliable with $ICC2K=0.771$, reflecting good consistency across participants.
We also investigate the relationship between response correctness and average response time, showing an inverse correlation ($Pearson=-0.284$).
This reveals that questions with higher accuracy tend to elicit shorter response times.

\paragraph{Overall Model Performance}
We present a comprehensive overview of the performance of all evaluated MLLMs in Figure \ref{fig:model performance}(a). 
This includes a direct comparison of accuracies achieved under both the single composite image and the separate images task settings.
The results highlight significant variability in performance, not only between different models but also across the two distinct evaluation paradigms. 
Closed-sourced models generally achieve higher accuracy than open-source models. 
Within open-source models, there is no significant performance difference based on model size; all open-sourced models perform comparably to a randomly sampled baseline. 
Furthermore, we investigate whether model response length correlates with accuracy, analogous to trends observed in human performance.
Using Gemini 2.5 Pro which provides token-level counts for both internal reasoning (``thinking'') and final response, we measure the Pearson correlation between response length and accuracy.
The resulting correlation coefficient is 0.021, indicating no meaningful relationship between the two and suggesting that, unlike in humans, longer responses do not reflect deeper or more accurate reasoning in the model.
A detailed breakdown of scores per model and per task category is provided in Table \ref{apptab:multiple images} and \ref{apptab: single image}.

\paragraph{Critique of Single Composite Image Evaluation}
Our findings indicate a notable discrepancy in model performance between the two evaluation settings in advanced models. 
Specifically, the single composite image approach consistently yielded lower accuracies by 4\% on average across GPT series models compared to the separate images setting. 
Most closed-source models significantly outperformed a random baseline ($p < 0.05$) when using separate images, whereas only GPT o3 and o4-mini showed significant difference from the baseline with a single composite image input. 
This observation suggests that the challenge in the single image setup may stem more from the complexities of parsing cluttered visual components and segregating distinct conceptual entities, rather than purely from a deficiency in spatial reasoning. 
Consequently, we posit that previous benchmarks employing solely this composite image methodology do not accurately reflect the intrinsic spatial cognition capabilities of current MLLMs. 
Therefore, we only discuss evaluation results with separate images as input in the following sections.

\rparagraph{Models are more likely to success on instances that humans perceive as easier}
We investigate whether MLLM performance is essentially random across different complexity levels reflected by human performance. 
Figure~\ref{fig:model performance}(b) plots model accuracy against average human accuracy for the same set of examples, revealing a general upward trend: models tend to perform better on instances that humans also find easier, indicated by positive slopes.
This correlation, supported by statistically significant tests against a random baseline, suggests that current MLLMs do exhibit early signs of spatial cognition.
While their reasoning remains limited, the non-random variation in performance across difficulty levels justifies the presence of spatial cognition in these models.

\subsection{Discussion and Analysis}
Building on our high-level performance analysis, we investigate whether instance-wise correctness can be predicted from relevant features. 
This is important for assessing the reliability of MLLMs: if consistent patterns exist, model correctness can be anticipated, enabling safer and more robust deployment \citep{zhou2023predictable}. 
Comparison with humans further reveals how closely MLLMs mirror human-like spatial reasoning and help to guide model development.

\paragraph{Analysis Setups}
To explore how well perceptual and reasoning features can explain behavior (cognitive profile), we use machine learning classifiers (random forest) to predict instance-level correctness for both humans and MLLMs.
To address label imbalance, class weights are adjusted inversely to class frequencies in the input data when training the classifier.
We consider two classification settings: binary classification (correct vs. incorrect) and four-class classification (0–3 correct responses across participants).
To further analyze cognitive effort, we apply linear regression to predict human response time and MLLM token counts including thinking using the same set of features.
The cognitive-related input features are introduced as follows, encompassing both perceptual and reasoning-related dimensions.

For visual perception, we include three features: the pattern complexity of both the question and the answer options, as well as the image resolution. 
Image resolution can impact perceptual recognition, with lower fidelity obscuring visual structure, so we discretize resolution into three bins (low, medium, high) to reflect practical perceptual clarity.
For general reasoning, we extract four features representing the number of reasoning steps required for each category of atomic operations: \textit{Pattern Matching}, \textit{Spatial Relation Analysis}, \textit{Spatial Manipulation}, and \textit{Logical Deduction}.
To ensure a stable signal from human, in addition to the correctness of individual human participant, we aggregated responses from three evaluators, as individual responses may be subject to idiosyncratic noise preventing reliable modeling of human cognitive profiles, while models are largely deterministic.

\begin{figure*}[t]
    \centering
    \includegraphics[trim={0cm 10.2cm 0cm 0cm}, clip, width=0.9\textwidth]{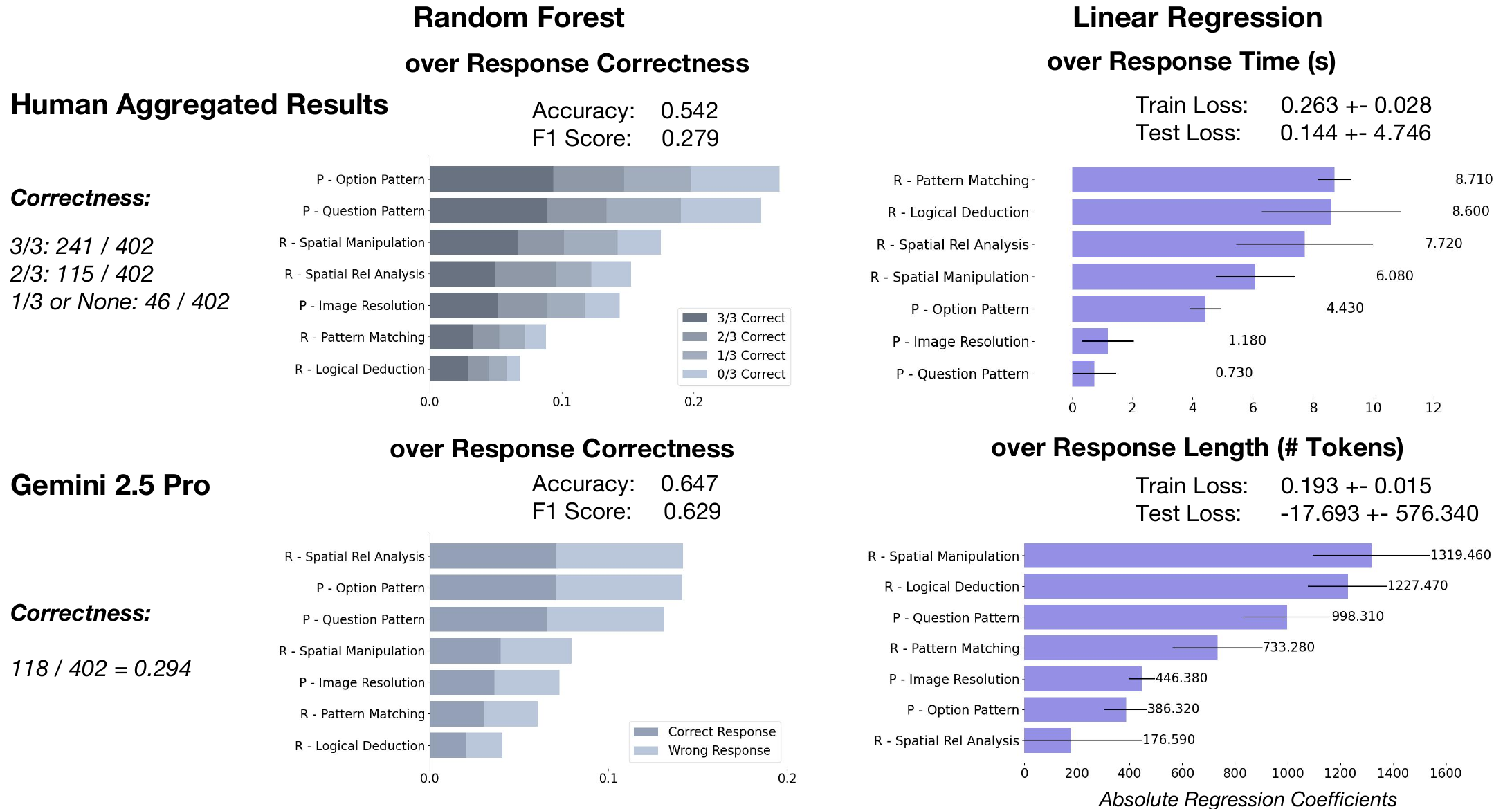}
    \caption{Cognitive profile analysis using SHAP values for correctness prediction and linear regression coefficients for cognitive load, comparing humans and MLLMs. More results in Figures~\ref{appfig:human cognitive pattern} and~\ref{appfig:model cognitive pattern}.}
    \label{fig:cognitive pattern}
\end{figure*}

\rparagraph{Human correctness is predictable while MLLMs exhibit near-random instance-level behavior}
We train the classifiers over the set of examples for human evaluation for fair comparison between human and models using 5-fold cross-validation. 
Our goal is not to maximize classification accuracy, but to identify the presence or absence of structured cognitive profiles.
To mitigate the effects of severe data imbalance and limited samples per fold due to high human accuracy and low model accuracy, we aggregate predictions across folds for more stable metric estimation.
Human correctness of individual participants is highly predictable with Random Forest, reaching weighted F1 scores of 0.631, 0.821 and 0.799 ($p<0.0002$) and AUC score of 0.579, 0.643 and 0.621. 
In the more granular four-class setting (aggregated human correctness), the classifier still performs above chance (F1 = 0.279 vs. 0.192, $p < 0.05$), reinforcing the presence of systematic cognitive behavior.
In contrast, classifiers trained on MLLM outputs fail to detect consistent correctness patterns. 
As shown in Figure~\ref{appfig:model cognitive pattern}, weighted F1 scores and AUC scores remain lower than human participants across most model variants, with no significant improvement over random baselines ($p>0.01$).
These results suggest that human responses are governed by predictable cognitive strategies, while current MLLMs lack the internal structure for reliable spatial reasoning at instance level.

\rparagraph{Pattern complexity drives human correctness, while reasoning features govern cognitive effort}
To understand which features contribute most to human success, we apply SHAP analysis to the trained classifiers.
As shown in Figures~\ref{fig:cognitive pattern} and~\ref{appfig:human cognitive pattern}, \textit{Pattern Complexity} (especially in answer options) is the strongest predictor of correctness across all participants.
This is followed by the presence of \textit{Spatial Manipulation}, a cognitively demanding reasoning step.
We further model human response time, a proxy for cognitive effort, using linear regression on the same features.
The model predicts time with average error <1 second (±4s), and analysis of coefficients shows that reasoning features (\textit{Spatial Relation Analysis}, \textit{Spatial Manipulation}, \textit{Logical Deduction}) are the dominant contributors to increased response time.
Interestingly, \textit{Pattern Matching} correlates with shorter response times, possibly due to heuristic strategies such as visual elimination or rule-of-thumb matching.
Together, these results highlight a dual cognitive profile in humans: while perceptual errors (e.g., misreading complex patterns) drive most mistakes, reasoning complexity governs cognitive effort.

\rparagraph{MLLMs show partial alignment with human profiles, but responses remain sensitive to low-level visual cues}
We apply SHAP analysis to the classifiers trained on MLLM correctness (Figure~\ref{appfig:model cognitive pattern}) and observe high variability across models, with most failing to reach statistical significance (denoted in \textcolor{orange}{\textit{orange}} with $p>0.01$).
Still, some convergence with human cognition emerges.
\textit{Option Pattern Complexity} is a shared influential feature across both humans and MLLMs, while features like \textit{Image Resolution} and \textit{Spatial Relation Analysis} are more prominent for certain MLLMs.
This suggests that while models do attend to meaningful patterns, they remain disproportionately influenced by low-level visual cues and spatial relationship understanding.
To further investigate MLLM effort, we model “thinking length” using linear regression.
Here, we find that in addition to reasoning-related features, \textit{Question Pattern Complexity} contributes significantly, while \textit{Spatial Relation Analysis} appears to be the least predictive factor, marking a clear divergence from human profiles.
These findings point to a hybrid picture: while MLLMs exhibit emerging spatial awareness, their instance-level reasoning remains noisy and constrained by understanding low-level visual cues, calling for further research.
\section{Conclusion}
This work introduced a novel framework with \benchmark benchmark for evaluating MLLMs' spatial cognition against human cognitive profiles, moving beyond aggregate accuracy with fine-grained analysis. 
Our findings show that while current MLLMs show early signs of spatial reasoning, their overall capabilities remain limited with randomness.
Human accuracy is consistently shaped by pattern complexity and reasoning demands, revealing structured and predictive cognitive profiles.
In contrast, model behavior is more influenced by understanding low-level visual cues such as image resolution and spatial relations, with less predictable and interpretable responses at instance level.
These results highlight both emerging capabilities and critical gaps between human and MLLMs spatial cognition. 
We hope our findings and \benchmark benchmark with finegrained cognitive feature annotations serve as a foundation for future research toward closing this gap, enabling the development of MLLMs with more robust, human-aligned spatial capabilities.

\bibliographystyle{neurips_2023}

\bibliography{custom}

\newpage
\appendix
\section{\benchmark}

\label{appsubsec:benchmark construction}

\paragraph{Overview of the Framework}
We introduce an evaluation framework designed for a fine-grained analysis of MLLMs' spatial reasoning capabilities. 
The framework extends beyond previous benchmarks in three crucial ways.

\textbf{\iparagraph{1. Disentangling Cognitive Features (\S\ref{subsec:confoundingfactors})}}
Previous benchmarks often conflate distinct cognitive features that affect model accuracy in spatial reasoning tasks, such as perceptual difficulty and inherent reasoning complexity.
Ignoring these features undermines evaluation validity and explainability, hindering real-world applicability when selecting appropriate models \citep{burden2023inferring}. 
Our framework explicitly identifies and accounts for these performance-affecting features:
\begin{itemize}[leftmargin=*]
\item \iparagraphnodot{Visual Perception}: Complex visual patterns require accurate interpretation of pattern structures before reasoning begins.
\item \iparagraphnodot{General Reasoning}: The inherent complexity of the reasoning process itself, e.g., requiring multiple reasoning hops or intricate spatial transformations, adds difficulty that might overshadow an MLLM's genuine spatial reasoning capabilities.
\end{itemize} 

\textbf{\iparagraph{2. Instance-Wise Evaluation with Predictive Power (\S\ref{subsec:discussion})}}
Typical average-based benchmark scores (e.g., accuracy) primarily represent overall performance, making it difficult to anticipate whether a model will correctly answer a new question. 
Inspired by \citet{zhou2025general}, our framework enhances interpretability by supporting instance-wise evaluation.
This allows researchers to estimate the likelihood that a model will correctly answer a given question based on known cognitive features \citep{burden2023inferring}, informing both deployment decisions and future research directions.

\textbf{\iparagraph{3. Parallel Analysis with Human Cognitive Profiles (\S\ref{subsec:discussion})}} 
Despite drawing inspiration from human cognitive tests, previous work lacks direct comparison with human cognition. 
We bridge this gap by incorporating human evaluation with \textit{response time} for each question as a proxy for human-perceived task difficulty \citep{barrouillet2007time,jintelligence4040014}. 
This parallel analysis reveals the extent to which current MLLMs emulate or diverge from human-like spatial cognition, offering insights to guide the advancement of MLLMs.

This dataset is for research purposes only and should not be used outside of research contexts. 

\paragraph{Data Source}
We construct the benchmark from two primary sources: a public subset collected from the web and a private subset sourced from purchased educational materials. 
For the public data, we crawl the web using carefully selected spatial reasoning keywords.
For the private dataset, we acquire spatial aptitude test materials from certified test preparation providers, targeting children under 11 years old.

To ensure the quality of the crawled data and retain only well-formed spatial problems, we implement a filtering pipeline that discard repetitive items based on the urls and ask human annotators to filter out samples that are irrelevant, ambiguous or do not evaluate spatial reasoning. 
All the data is expressed in English.

\paragraph{Targeted Capabilities and Task Types}
We focus on spatial cognition tasks designed for young adolescents, using the 11+ exam level as an anchor. 
Given that not all spatial cognitive skills are equally suited for evaluation in MLLMs, we concentrate on the following three core capabilities: \textit{Spatial Relation and Orientation}, \textit{Spatial Visualization} and \textit{Flexibility of Closure}. 
Each capability encompasses a collection of tasks, with definitions and examples summarized in Table~\ref{apptab:capability and task}. 
The selected tasks emphasize interpretable reasoning steps and perceptual challenges amenable to MLLM analysis.

\newcolumntype{L}{>{\raggedright\arraybackslash}X}

\begin{table}[htbp]
\centering
\caption{Spatial capabilities and corresponding tasks, with question descriptions and number of examples in public and private split. }
\label{tab:my_robust_table}
\begin{tabularx}{\linewidth}{@{} p{2.3cm} p{1.8cm} L c c @{}} 
\toprule
\textbf{Capability} & \textbf{Task} & \textbf{Question Description} & \textbf{Public} & \textbf{Private} \\ 
\midrule
Spatial Relation and Orientation & 2D shape rotation (SRO.1) & The image shows several 2D shapes, including a designated target shape. Select the option that is the target shape rotated to a different orientation. & 35 & 10 \\ 
\cmidrule(l){2-5}
& 2D shape reflection (SRO.2) & The image displays several 2D shapes, with one identified as the target shape. The target shape has been reflected across a mirror line shown in the image. & 33 & - \\ 
\cmidrule(l){2-5}
& 3D shape rotation (SRO.3) & This image shows a 3D polycube shape. Choose the option that represents the same shape, viewed from a different rotation. & 6 & 3 \\ 
\midrule
Spatial Visualization & Shape completion (SV.1) & The image presents an equation involving a target shape and several shape candidates that can be added to or removed from the base shape. & 9 & 10 \\ 
\cmidrule(l){2-5}
& Shape combination (SV.2) & The image illustrates an equation involving a basic shape, where shapes are either added or removed. Only edges labeled with the same letter can be combined. & 68 & 10 \\ 
\cmidrule(l){2-5}
& Building blocks (SV.3) & The image displays a target complex 3D shape along with several sets of blocks. Identify the set of blocks that can be combined to form the target shape. & 52 & 10 \\ 
\cmidrule(l){2-5}
& Paper folding (SV.4) & The image shows a piece of paper being folded and then punched with holes. Select the option that correctly shows the pattern of holes after the paper is fully unfolded. & 229 & 9 \\ 
\cmidrule(l){2-5}
& Cube and nets (SV.5) & The image shows an unfolded shape (net) and several cube candidates. Identify which option can be correctly folded into a cube from the given unfolded shape. & 201 & 9 \\ 
\midrule
Flexibility of Closure & Hidden shape (FoC) & The target shape is hidden within one of the answer options. It may be rotated and embedded within the option. Identify the option that contains the hidden target shape. & 76 & 10 \\ 
\midrule
Comprehensive (SV+SRO) & Cube and dice (Com.1) & The image shows different views of the same cube, with a unique symbol on each of its six faces. Determine which option correctly matches the missing face. & 17 & 10 \\
\cmidrule(l){2-5}
& 3D-2D view (Com.2) & This image displays a 3D object. Select the option that correctly represents a 2D view of the object from a specific perspective. & 98 & 10 \\
\bottomrule
\end{tabularx}
\label{apptab:capability and task}
\end{table}

\paragraph{Expert Annotation Protocol}
We recruit three domain experts to annotate the benchmark data. All annotators hold postgraduate degrees or higher in STEM fields, with backgrounds in mathematics or engineering. The annotation process adheres to institutional ethical guidelines. 
All annotations are collected anonymously and no information that names or uniquely identifies individual people or offensive content are collected or used.
The instructions explain that the data would be used for research purpose only.

\paragraph{Annotation Fields and Guidelines}
As described in Section~\ref{subsec:confoundingfactors}, all samples are annotated for two cognitive dimensions: \textbf{\textit{Visual Perception Complexity}} and \textbf{\textit{General Reasoning Process}}.

For tasks with highly standardized visual transformations, such as 2D shape rotation, 2D shape reflection, or 3D-2D view, we do not require annotators to document full reasoning steps, as these processes are straightforward and consistent across samples. 
For all other tasks, each expert independently provides both visual perception and reasoning annotations according to the detailed protocol described below.

\textbf{\textit{Visual Perception Complexity~}}
We quantify visual complexity for both the question and the option choices. 
The complexity score is derived from the number of atomic components in each pattern. 
We define atomic components as key features:
\begin{itemize}[leftmargin=*]
    \item For referable shapes (e.g., heart, star), complexity is based on the number of symbolic elements.
    \item For abstract 2D patterns, we count the number of lines or segments.
    \item For abstract 3D structures, we count the number of surfaces or faces.
\end{itemize}
This methodology yields a consistent, interpretable complexity score for each visual input. Example annotations are shown in Figure~\ref{fig:data collection}.

\textbf{\textit{General Reasoning Process~}}
To capture the reasoning process, we define a taxonomy of atomic operations that cover a wide range of spatial reasoning strategies. 
Annotators must select one operation per step from the categories defined below:

\textit{Pattern Matching}: Determine whether one entity visually contains or resembles another.
The match can be based on exact visual similarity or shared key features. Shape matching does not involve reasoning about spatial relationships, nor does it alter the spatial properties of the entities involved.

\begin{verbatim}
def pattern_match(entity_a: Object, entity_b: Object) -> bool
\end{verbatim}

\textit{Spatial Relation Analysis}: Analyze the spatial relationship between two entities.
Any two non-overlapping 2D or 3D shapes can be treated as separate entities, for example, two cubes, or two faces of the same cube, depending on the context of analysis. 
This process does not change the spatial properties or the overall spatial layout of the entities.
Subtypes include:
\begin{itemize}
    \item Position: Determine the relative position of shape B within entity A.
    \item Orientation: Determine the direction a part of shape A or entity B is facing (e.g., "Part X of A points toward C").
    \item Perspective: Infer the viewpoint (e.g., "viewed from behind").
    \item Rotation: Determine the direction or angle of rotation.
    \item Folding: Determine the direction in which a 2D net folds into a 3D object.
    \item Projection: Determine the direction in which a 3D entity is projected onto a 2D plane.
\end{itemize}
\begin{verbatim}
def spatial_relation(entity_a: Object, entity_b: Object) -> statement: str
\end{verbatim}

\textit{Spatial Manipulation}: Change the spatial properties or overall spatial layout of entities.
\begin{itemize}
    \item 2D operations: rotation, translation, reflection, adding/removing shapes
    \item 3D operations: 3D rotation (around an axis), 3D translation, 3D symmetry
    \item Dimensional transformations: projection in a certain direction, folding along an edge
    \item Counting: e.g., counting the number of holes in an origami structure
    \item Symbol tagging: labeling shapes or parts with markers or symbols
\end{itemize}
\begin{verbatim}
def spatial_manipulate(entity: Object, statement: str) -> Union[Object, str]
\end{verbatim}

\textit{Logical Deduction}: Infer rules or verify spatial conditions.
\begin{itemize}
    \item Logical inference: inferring spatial properties or rules, such as:
    \begin{itemize}
        \item "A cannot be adjacent to B"
        \item "A must be opposite to C"
        \item "Cube A can be obtained from Cube B via one or two rotations"
    \end{itemize}
    \item Verification: testing whether a property or rule holds on another entity
\end{itemize}
\begin{verbatim}
def logical_deduction(*statements: str) -> Union[str, bool]
\end{verbatim}

Annotators are instructed to decompose their reasoning into step-by-step sequences using these operations, ensuring consistency and reproducibility. 
This structured representation enables us to map human reasoning steps to potential model behaviors.

\begin{figure*}[t]
    \centering
    \includegraphics[trim={0cm 18cm 5cm 0cm}, clip, width=0.8\textwidth]{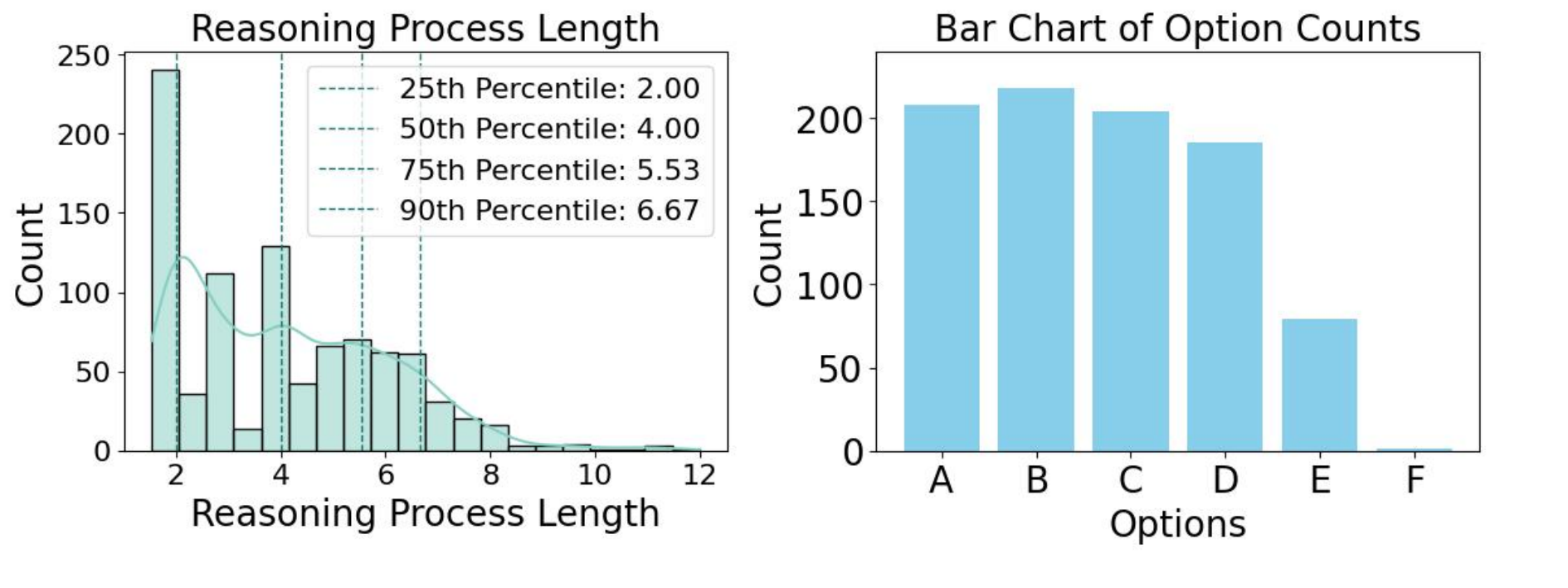}
    \caption{Data distributions over lengths of reasoning process and golden options. }
    \label{appfig:data distribution}
\end{figure*}
\section{Experiments}

\subsection{Human Participants}
We recruit three human participants as evaluators to evaluate human performance and record human behavior (response time in seconds). 
They are not involved in the annotation process with STEM major background for bachelors major, such as Informatics and AI. 
All the human evaluators are gathered physically to conduct human evaluations, making sure that the performance really reflects their abilities and behaviors.

\subsection{Models}
\paragraph{Hyperparameters}
We adopt most of the inference parameters by default for proprietary models. 
For open-sourced models, we adopt the default configuration in HuggingFace. 

\paragraph{Prompts}
Table \ref{apptab:prompt single} and \ref{apptab:prompt multiple} show the prompt templates for single image setting and separate image setting respectively. 
Within the prompt templates, <QUESTION> and <OPTIONS> are replaced with the questions in Table \ref{apptab:capability and task} for different tasks. 

\begin{table*}[]
\begin{tcolorbox}[title = {Single Image Input}]
<QUESTION> 

<image>

Conclude your chosen answer to the multiple-choice question between <ANSWER> and </ANSWER>. 
\end{tcolorbox}
\caption{Prompt templates for main experiments with single image as input. }
\label{apptab:prompt single}
\end{table*}

\begin{table*}[]
\begin{tcolorbox}[title = {Separate Image Input}]
<QUESTION>

<image>

A: 
<image>

B: 
<image>

C: 
<image>

D: 
<image>

E: 
<image>

Conclude your chosen answer to the multiple-choice question between <ANSWER> and </ANSWER>. 
\end{tcolorbox}
\caption{Prompt templates for main experiments with separate image segments as input. }
\label{apptab:prompt multiple}
\end{table*}

\subsection{Results}

A detailed breakdown of scores per model and per task category is provided in Table \ref{apptab:multiple images} and \ref{apptab: single image} for multiple separate images and single image as input.

\begin{table}
\centering
\renewcommand\arraystretch{1.4}
\caption{Task-wise performance per model with separate multiple images as input. }
\resizebox{\linewidth}{!}{%
\begin{tabular}{l|ccccccccccc} 
\hline
Model                    & SRO.1 & SRO.2 & SRO.3 & SV.1 & SV.2 & SV.3 & SV.4 & SV.5 & FoC & Com.1 & Com.2  \\ 
\hline
GPT 4o                   & 0.267             & 0.485               & 0.444             & 0.158            & 0.128             & 0.290           & 0.357         & 0.257         & 0.279        & 0.222         & 0.370       \\
GPT 4.1-mini             & 0.289             & 0.273               & 0.333             & 0.368            & 0.295             & 0.194           & 0.340         & 0.248         & 0.279        & 0.074         & 0.278       \\
GPT 4.1-nano             & 0.200             & 0.394               & 0.444             & 0.211            & 0.192             & 0.387           & 0.269         & 0.195         & 0.163        & 0.185         & 0.269       \\
GPT-o1                   & 0.378             & 0.364               & 0.444             & 0.158            & 0.205             & 0.258           & 0.445         & 0.338         & 0.256        & 0.222         & 0.324       \\
GPT-o3                   & 0.444             & 0.485               & 0.556             & 0.316            & 0.295             & 0.274           & 0.458         & 0.448         & 0.349        & 0.185         & 0.306       \\
GPT-o4-mini              & 0.267             & 0.485               & 0.444             & 0.263            & 0.231             & 0.452           & 0.395         & 0.305         & 0.349        & 0.185         & 0.306       \\
Gemini 2.0 Flash         & 0.222             & 0.212               & 0.444             & 0.158            & 0.179             & 0.323           & 0.382         & 0.257         & 0.267        & 0.185         & 0.278       \\
Gemini 2.5 Flash & 0.356             & 0.242               & 0.444             & 0.211            & 0.269             & 0.339           & 0.395         & 0.276         & 0.174        & 0.296         & 0.315       \\
Gemini 2.5 Pro           & 0.333             & 0.394               & 0.222             & 0.263            & 0.308             & 0.323           & 0.378         & 0.300         & 0.128        & 0.296         & 0.324       \\ 
\hline
\multicolumn{12}{c}{Open-Sourced Models}                                                                                                                                                                                                        \\ 
\hline
Qwen 2.5VL 3B            & 0.267             & 0.182               & 0.333             & 0.158            & 0.295             & 0.387           & 0.235         & 0.276         & 0.198        & 0.259         & 0.278       \\
Qwen 2.5VL 7B            & 0.133             & 0.424               & 0.111             & 0.211            & 0.218             & 0.387           & 0.218         & 0.214         & 0.209        & 0.407         & 0.306       \\
Gemma3 12B               & 0.289             & 0.212               & 0.333             & 0.316            & 0.154             & 0.242           & 0.265         & 0.205         & 0.209        & 0.185         & 0.157       \\
Gemma3 27B               & 0.178             & 0.303               & 0.333             & 0.211            & 0.192             & 0.258           & 0.324         & 0.238         & 0.128        & 0.259         & 0.231       \\
\hline
\end{tabular}
}
\label{apptab:multiple images}
\end{table}

\begin{table}
\centering
\renewcommand\arraystretch{1.4}
\caption{Task-wise performance per model with single images as input. }
\resizebox{\linewidth}{!}{%
\begin{tabular}{l|ccccccccccc} 
\hline
Model                    & SRO.1 & SRO.2 & SRO.3 & SV.1 & SV.2 & SV.3 & SV.4 & SV.5 & FoC & Com.1 & Com.2  \\ 
\hline
GPT 4o                   & 0.156             & 0.364               & 0.333             & 0.368            & 0.256             & 0.371           & 0.248         & 0.224         & 0.244        & 0.407         & 0.231       \\
GPT 4.1-mini             & 0.111             & 0.242               & 0.556             & 0.211            & 0.167             & 0.290           & 0.265         & 0.214         & 0.291        & 0.185         & 0.250       \\
GPT 4.1-nano             & 0.267             & 0.303               & 0.556             & 0.105            & 0.218             & 0.323           & 0.311         & 0.186         & 0.221        & 0.185         & 0.130       \\
GPT-o1                   & 0.200             & 0.364               & 0.444             & 0.211            & 0.167             & 0.242           & 0.261         & 0.238         & 0.233        & 0.185         & 0.250       \\
GPT-o3                   & 0.378             & 0.273               & 0.444             & 0.158            & 0.282             & 0.306           & 0.382         & 0.400         & 0.221        & 0.148         & 0.231       \\
GPT-o4-mini              & 0.311             & 0.394               & 0.222             & 0.211            & 0.218             & 0.339           & 0.332         & 0.300         & 0.291        & 0.148         & 0.278       \\
Gemini 2.0 Flash         & 0.178             & 0.242               & 0.333             & 0.211            & 0.128             & 0.435           & 0.298         & 0.276         & 0.256        & 0.111         & 0.204       \\
Gemini 2.5 Flash & 0.178             & 0.242               & 0.333             & 0.211            & 0.128             & 0.435           & 0.298         & 0.276         & 0.256        & 0.111         & 0.204       \\
Gemini 2.5 Pro           & 0.267             & 0.333               & 0.333             & 0.263            & 0.205             & 0.323           & 0.387         & 0.410         & 0.279        & 0.296         & 0.259       \\ 
\hline
\multicolumn{12}{c}{Open-Sourced Models}                                                                                                                                                                                     \\ 
\hline
Qwen 2.5VL 3B            & 0.267             & 0.212               & 0.222             & 0.211            & 0.269             & 0.194           & 0.227         & 0.276         & 0.279        & 0.185         & 0.176       \\
Qwen 2.5VL 7B            & 0.333             & 0.212               & 0.111             & 0.211            & 0.321             & 0.435           & 0.235         & 0.229         & 0.174        & 0.111         & 0.250       \\
Gemma3 12B               & 0.156             & 0.212               & 0.222             & 0.316            & 0.282             & 0.371           & 0.231         & 0.229         & 0.256        & 0.370         & 0.241       \\
Gemma3 27B               & 0.200             & 0.212               & 0.111             & 0.211            & 0.244             & 0.274           & 0.227         & 0.224         & 0.221        & 0.111         & 0.139       \\
\hline
\end{tabular}
}
\label{apptab: single image}
\end{table}

\begin{figure*}[t]
    \centering
    \includegraphics[trim={0cm 0cm 0cm 0cm}, clip, width=\textwidth]{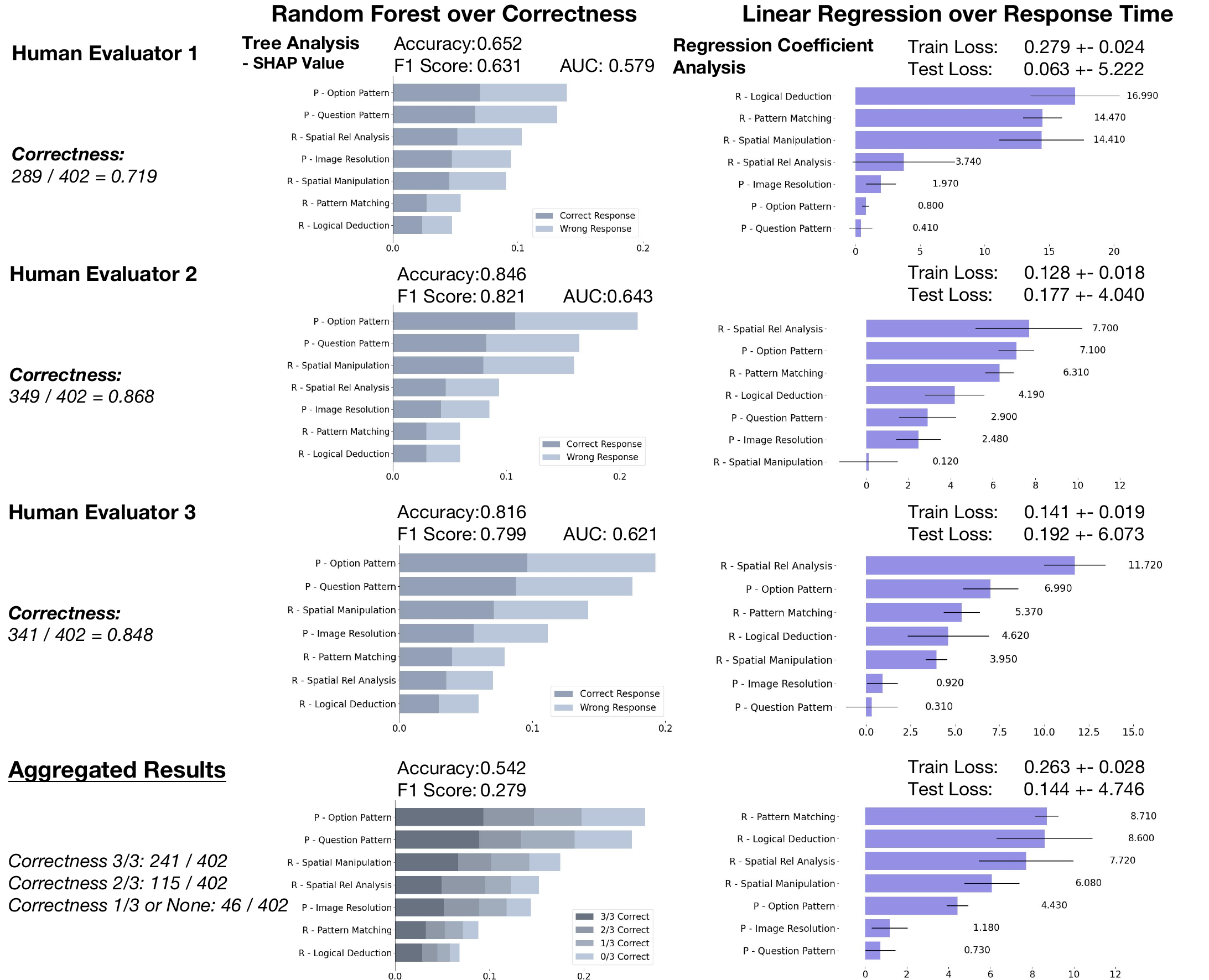}
    \caption{Feature Relevance in the Cognitive Profiles of Individual Human Participants and Aggregated Human Behavior. Individual human responses are predictable with $p<0.0002$ for F1 score compared to random chance. }
    \label{appfig:human cognitive pattern}
\end{figure*}

\begin{figure*}[t]
    \centering
    \includegraphics[trim={0cm 2.5cm 0cm 0cm}, clip, width=\textwidth]{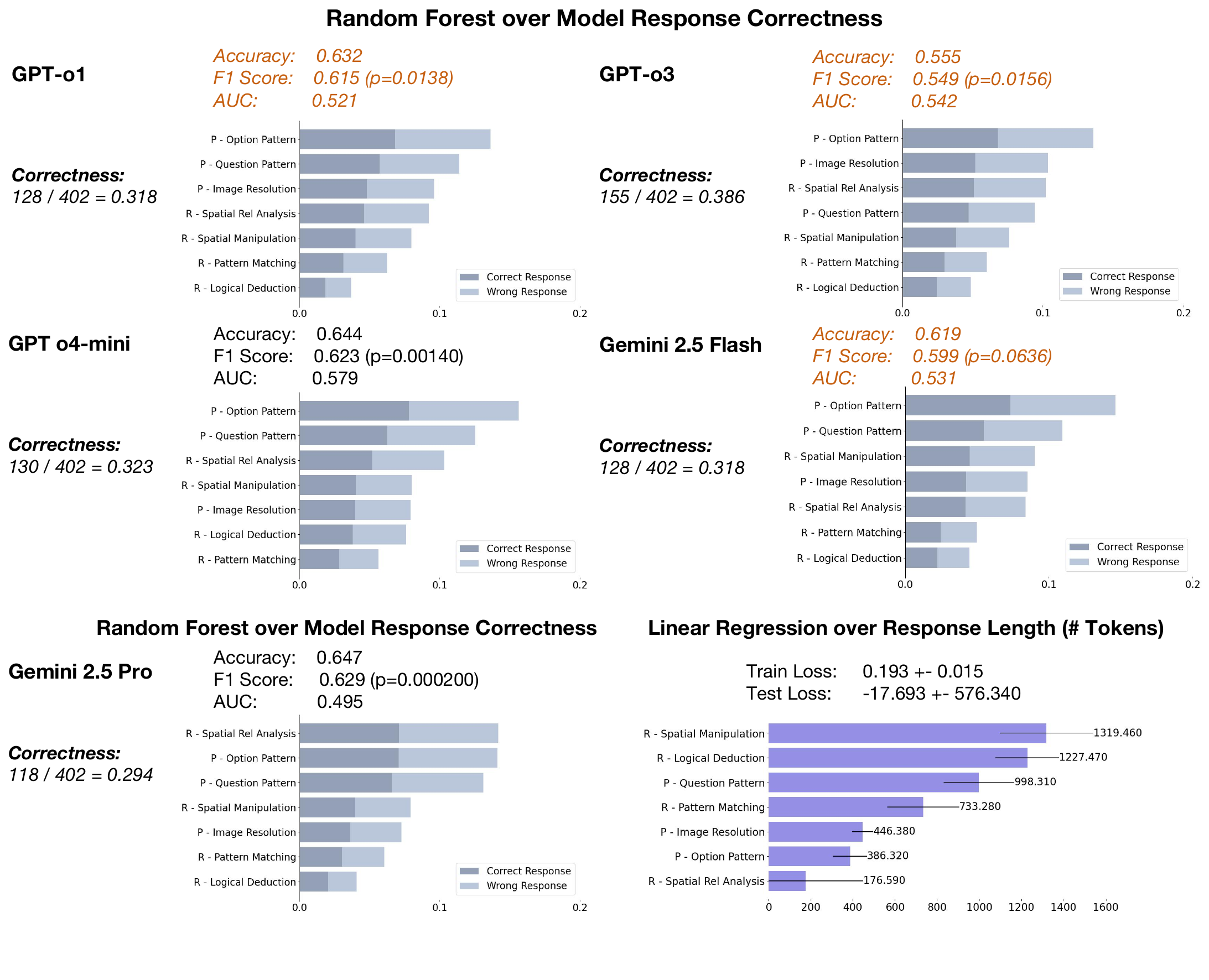}
    \caption{Feature Relevance in the Cognitive Profiles of Different Model Variants. }
    \label{appfig:model cognitive pattern}
\end{figure*}

Figure~\ref{appfig:human cognitive pattern} and~\ref{appfig:model cognitive pattern} present extended cognitive pattern analyses across individual human participants and a broader set of MLLM variants.
For human participants, \textit{Pattern Complexity} consistently ranks as the most influential factor for correctness, while \textit{Logical Deduction} and \textit{Pattern Matching} appear less impactful.
Moreover, reasoning-related features contribute most significantly to response time, whereas perceptual features such as \textit{Pattern Complexity} and \textit{Image Resolution} are among the least influential in determining response time per sample.

In contrast, classifiers trained to predict MLLM correctness do not significantly outperform a random baseline, as indicated by the orange highlights in Figure~\ref{appfig:model cognitive pattern}.
No consistent cognitive profiles emerge across model variants: different features dominate in different models, suggesting a lack of stable, interpretable reasoning strategies in current MLLMs.

\begin{table*}
\renewcommand{\arraystretch}{1.1}
\caption{Response format parsing result with single image as input.}
\centering
\resizebox{\linewidth}{!}{
    \newcolumntype{R}{>{\centering\arraybackslash}p{0.13\linewidth}} %
    \newcolumntype{T}{>{\centering\arraybackslash}p{0.08\linewidth}}
\begin{tabular}{lTTTTTRR} 
\hline
\multirow{2}{*}{Model}   
 & \multirow{2}{*}{Success}         
 & \multirow{2}{*}{Ordinal}  & \multirow{2}{*}{Number }  & \multirow{2}{*}{Letter }  
 & \multirow{2}{*}{Unknown}  & Verbalized & Parsing  \\ 
 & & & & & &  Choice &  Failure \\
\hline
GPT 4o                   & 843     & 10      & 34     & -      & 27                     & -                 & 1                \\
GPT 4.1-mini             & 846     & 14      & 43     & 3      & 9                      & -                 & -                \\
GPT 4.1-nano             & 801     & 3       & 50     & 16     & 25                     & 20                &                  \\
GPT-o1                   & 852     & -       & 31     & 3      & 27                     & 1                 & 1                \\
GPT-o3                   & 862     & 1       & 34     & 1      & 16                     & -                 & 1                \\
GPT-o4-mini              & 818     & -       & 33     & 5      & 21                     & 34                & 3                \\
GPT 4.1                  & 855     & 6       & 30     & -      & 24                     & -                 & -                \\
Gemini 2.0 Flash         & 728     & 12      & 19     & 4      & 23                     & 129               & -                \\
Gemini 2.5 Flash &         &         &        &        &                        &                   &                  \\
Gemini 2.5 Pro           &         &         &        &        &                        &                   &                  \\
Qwen 2.5VL 3B            & 814     & -       & 22     & 15     & 48                     & 13                & 3                \\
Qwen 2.5VL 7B            & 829     & -       & 26     & 18     & 39                     & 3                 & -                \\
Gemma3 12B               & 824     & 1       & 55     & 1      & 11                     & 22                & 1                \\
Gemma3 27B               & 817     & 5       & 44     & 10     & 37                     & 1                 & 1                \\
\hline
\end{tabular}
}
\end{table*}

\begin{table*}
\renewcommand{\arraystretch}{1.1}
\caption{Response format parsing result with separate multiple images as inputs.}
\centering
\resizebox{\linewidth}{!}{
    \newcolumntype{R}{>{\centering\arraybackslash}p{0.13\linewidth}} %
    \newcolumntype{T}{>{\centering\arraybackslash}p{0.08\linewidth}}
\begin{tabular}{lTTTTTRR} 
\hline
\multirow{2}{*}{Model}   
 & \multirow{2}{*}{Success}         
 & \multirow{2}{*}{Ordinal}  & \multirow{2}{*}{Number }  & \multirow{2}{*}{Letter }  
 & \multirow{2}{*}{Unknown}  & Verbalized & Parsing  \\ 
 & & & & & &  Choice &  Failure \\
\hline
GPT 4o                   & 901     & -       & 7      & -      & 2       & 1                 & 4                \\
GPT 4.1-mini             & 900     & -       & 12     & 2      & -       & -                 & 1                \\
GPT 4.1-nano             & 866     & -       & 18     & 13     & 6       & 12                & -                \\
GPT-o1                   & 903     & -       & 7      & 2      & 2       & -                 & 1                \\
GPT-o3                   & 910     & -       & 2      & 1      & -       & 1                 & 1                \\
GPT-o4-mini              & 863     & -       & 11     & 1      & -       & 38                & 2                \\
GPT 4.1                  & 898     & -       & 13     & 1      & 3       & -                 & -                \\
Gemini 2.0 Flash         & 642     & -       & -      & -      & -       & 273               & -                \\
Gemini 2.5 Flash & -       & -       & -      & -      & -       & 909               & 5                \\
Gemini 2.5 Pro           &         &         &        &        &         &                   &                  \\
Qwen 2.5VL 3B            & 752     & -       & 5      & 11     & 28      & 119               & -                \\
Qwen 2.5VL 7B            & 848     & -       & 5      & 39     & 22      & 1                 & -                \\
Gemma3 12B               & 881     & -       & 1      & 3      & -       & 28                & 2                \\
Gemma3 27B               & 903     & -       & 6      & -      & 2       & 4                 & -                \\
\hline
\end{tabular}
}
\end{table*}

\end{document}